\documentclass[final]{./aux/cvpr}

\usepackage{times}
\usepackage{epsfig}
\usepackage{graphicx}
\usepackage{amsmath}
\usepackage{amssymb}
\usepackage{subfigure}

\usepackage[pagebackref=true,breaklinks=true,colorlinks,bookmarks=false]{hyperref}

\def\cvprPaperID{****} 
\def\confYear{CVPR 2021}

\begin{document}

\title{The Herbarium 2021 Half--Earth Challenge Dataset}

\author{Riccardo de Lutio\\
ETH Z\"urich \\
{\tt\small rdelutio@ethz.ch}
\and
Damon Little\\
New York Botanical Garden
\and
Barbara Ambrose\\
New York Botanical Garden
\and
Serge Belongie\\
Cornell Tech \& Google Research
}

\maketitle

\global\csname @topnum\endcsname 0
\global\csname @botnum\endcsname 0
\begin{abstract}

Herbarium sheets present a unique view of the world's botanical history, evolution, and diversity. This makes them an all--important data source for botanical research. With the increased digitisation of herbaria worldwide and the advances in the fine--grained classification domain that can facilitate automatic identification of herbarium specimens, there are a lot of opportunities for supporting research in this field. However, existing datasets are either too small, or not diverse enough, in terms of represented taxa, geographic distribution or host institutions. Furthermore, aggregating multiple datasets is difficult as taxa exist under a multitude of different names and the taxonomy requires alignment to a common reference. 
We present the Herbarium Half--Earth dataset, the largest and most diverse dataset of herbarium specimens to date for automatic taxon recognition.
\end{abstract}

\section{Introduction}
\begin{figure}[t]
\begin{center}
\includegraphics[trim=1cm 4cm 3cm 4cm, width=0.95\linewidth]{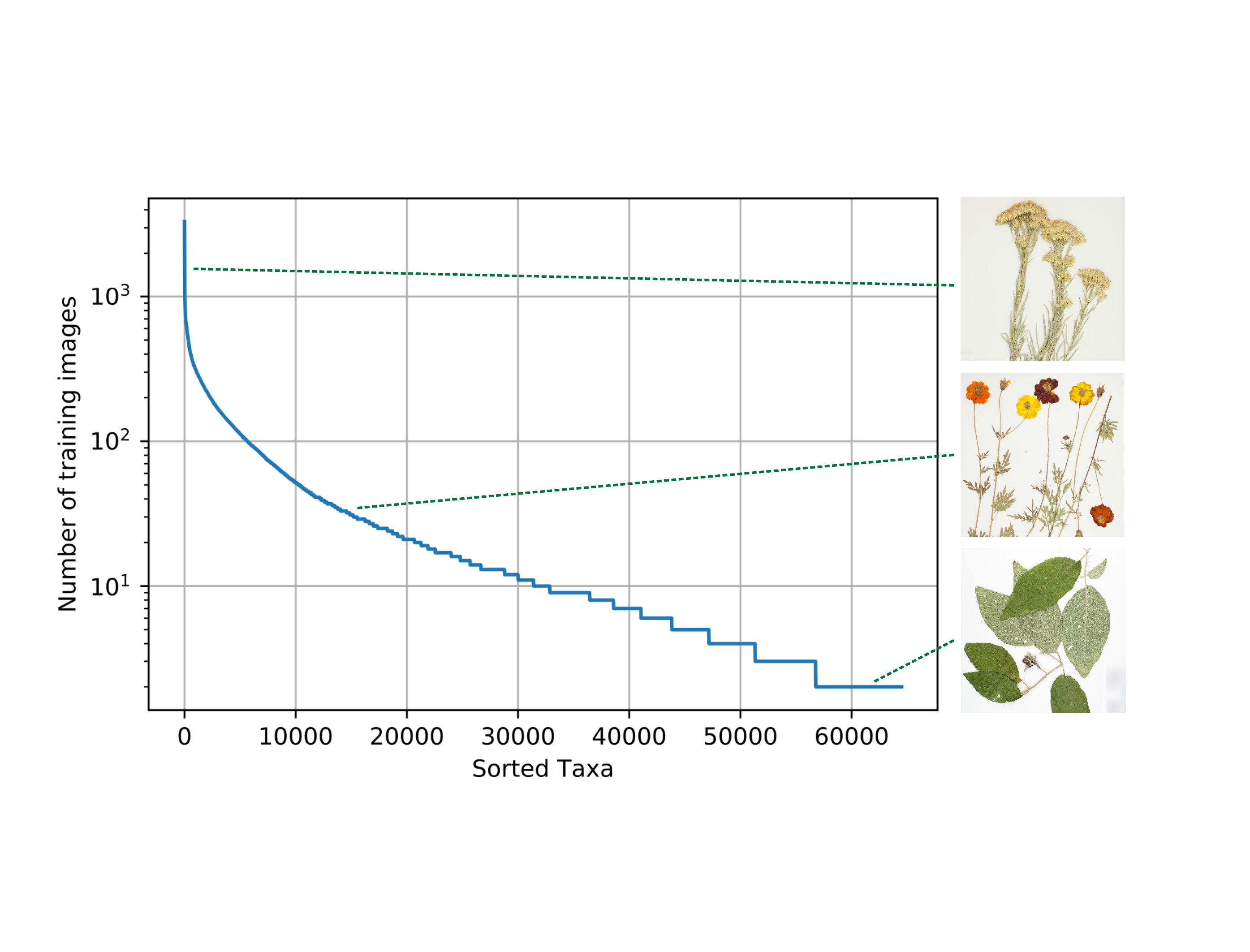}
\end{center}
\caption{Distribution of training images per taxon. The Herbarium Half--Earth dataset is highly imbalanced. Featured taxa are from top to bottom: \textit{Ericameria nauseosa} (Pall.~ex Pursh) G.L.~Nesom \& G.I.~Baird (Asteraceae), \textit{Bidens sulphurea} (Cav.)~Sch.~Bip.~(Asteraceae) and \textit{Solanum rixosum} A.R.~Bean (Solanaceae).}
\label{fig:dist}
\end{figure}

Natural history collections, such as herbarium specimens, contain a plethora of information from phenotype to genotype. Each specimen is a snapshot in time and all together provide a history of plants on Earth since the first herbarium collections were made nearly 500 years ago \cite{entibi}. Therefore, herbarium specimens are integral for understanding biodiversity and providing data to ameliorate the impacts of habitat loss and climate change \cite{calinger2013,davis2015,lang2019}.

Citizen science initiatives such as iNaturalist \cite{inat2017} and Pl@ntNet \cite{joly2016}, have popularised species recognition as a challenging real--world classification task, with large imbalanced fine--grained datasets. Similarly using computer vision methods for the automatic classification of herbarium specimens is a well studied topic, \cite{carranzarojas2017,clark2012,dillen2018,herbarium_journal,lorieul2019,pearson2020,ubbens2017,unger2016,waldchen2018,wijesingha2012,wilf2016,younis2018}. 
Many of these works focus on morphological trait recognition \cite{clark2012,lorieul2019,pearson2020,PryerEtAl2020,ubbens2017,unger2016,younis2018}, while others focus on species recognition from leaves only \cite{unger2016,wijesingha2012,wilf2016}. 

However existing datasets designed for computer vision approaches currently present some limitations. They are either small, targeted at specific taxa, only representative of a certain geographic region or coming from a single institution (see Tab. \ref{tab:datasets}). 
With the Herbarium Half--Earth dataset, we aim to address all these limitations and present the largest and most diverse dataset of herbarium specimens for automatic taxon recognition to date.

\begin{table}[ht]
\begin{center}
\resizebox{0.5\textwidth}{!}{%
\begin{tabular}{|l|c|c|c|c|c|}
\hline
\textbf{Dataset} & \textbf{\# Images} & \textbf{\# Taxa} & \textbf{\# Institutions} & \textbf{Geo. Range} \\
\hline
Dillen et al. \cite{dillen2018} & 1'900 & 1'580 & 9 & All Continents \\
\hline
Lorieul et al. \cite{lorieul2019} & 163'233 & 7'782 & 1 & Americas \\
\hline
Herbarium 255 \cite{carranzarojas2017} & 11'071 & 255 & 1 & Costa Rica \\
\hline
Herbarium 1K \cite{carranzarojas2017} & 253'733 & 1'204 & 1 & France\\
\hline
Herbarium 2019 \cite{herbarium2019} & 46'000 & 680 & 1 & Americas\\
\hline
Herbarium 2020 & 1'170'000 & 32'000 & 1 & Americas \\
\hline
\textbf{Herbarium 2021} & 2'500'000 & 64'500 & 5 & Americas, Oceania and Pacific \\
\hline
\end{tabular}
}
\end{center}
\caption{Summary of existing herbarium sheet datasets. Note that the Herbarium 2019 dataset focuses on the flowering plant family Melastomataceae, while the other datasets present a wider taxonomic diversity.}
\label{tab:datasets}
\end{table}

\section{The Herbarium Half--Earth Dataset}

The Herbarium Half--Earth dataset \footnote{\url{https://github.com/visipedia/herbarium_comp}} includes more than 2.5M images of vascular plant specimens representing nearly 64,500 taxa from the Americas and Oceania. 

The most exact labels are, in many cases, intraspecific (subspecies, varieties, forms, etc.)~or nothospecies (hybrids) neither of which can be characterized as ``species'', thus we use the terms ``taxon'' and ``taxa'' as generic descriptors of taxonomic labels.
In addition to labels for species--level and below, we also include labels at higher levels in the taxonomic hierarchy: family and order. This allows for experimentation with methods that address label hierarchy and label similarity.
These labels may also be supplemented by more fine--grained estimates of difference among taxa available from other sources \cite{JinQian2019}.

The images are provided by the New York Botanical Garden (NY), Bishop Museum (BPBM), Naturalis Biodiversity Center (NL), Queensland Herbarium (BRI), and Auckland War Memorial Museum (AK).

\begin{figure}[h]
\begin{center}
\includegraphics[width=0.95\linewidth]{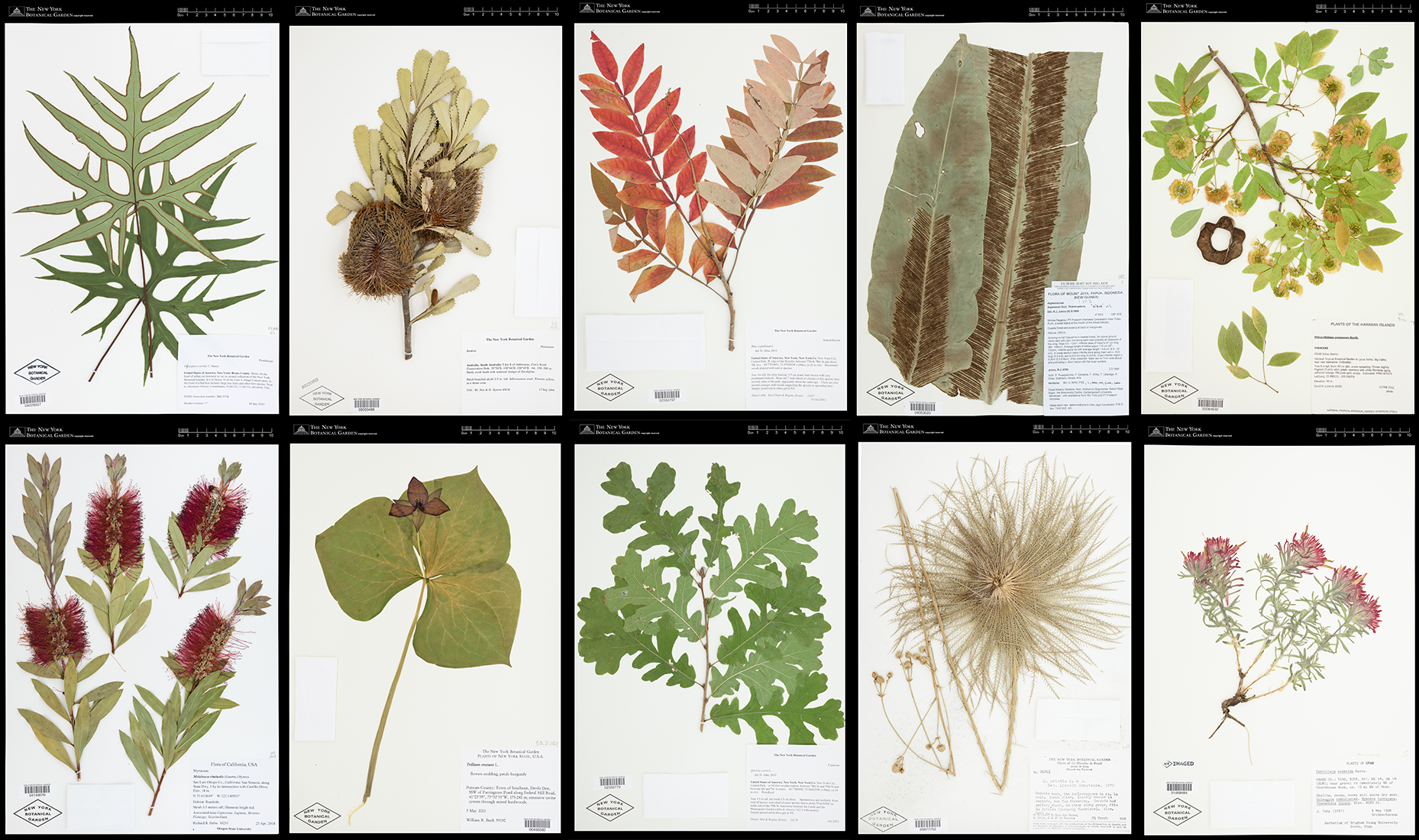}
\end{center}
   \caption{Example of images in the Herbarium Half--Earth dataset.}
\label{fig:ex}
\end{figure}

This dataset has a long tail; there are a minimum of three images per taxon (Fig.~\ref{fig:dist}). However, some taxa can be represented by more than 100 images. This dataset only includes images of vascular plant---the group of plants that includes lycophytes, ferns, gymnosperms, and flowering plants (Fig.~\ref{fig:ex}). The extinct forms of lycophytes are the major component of coal deposits, ferns are indicators of ecosystem health, gymnosperms provide major habitats for animals, and flowering plants provide almost all of our crops, vegetables, and fruits.

\subsection{Dataset Challenges}

\begin{figure}[h]
\begin{center}
\includegraphics[width=0.95\linewidth]{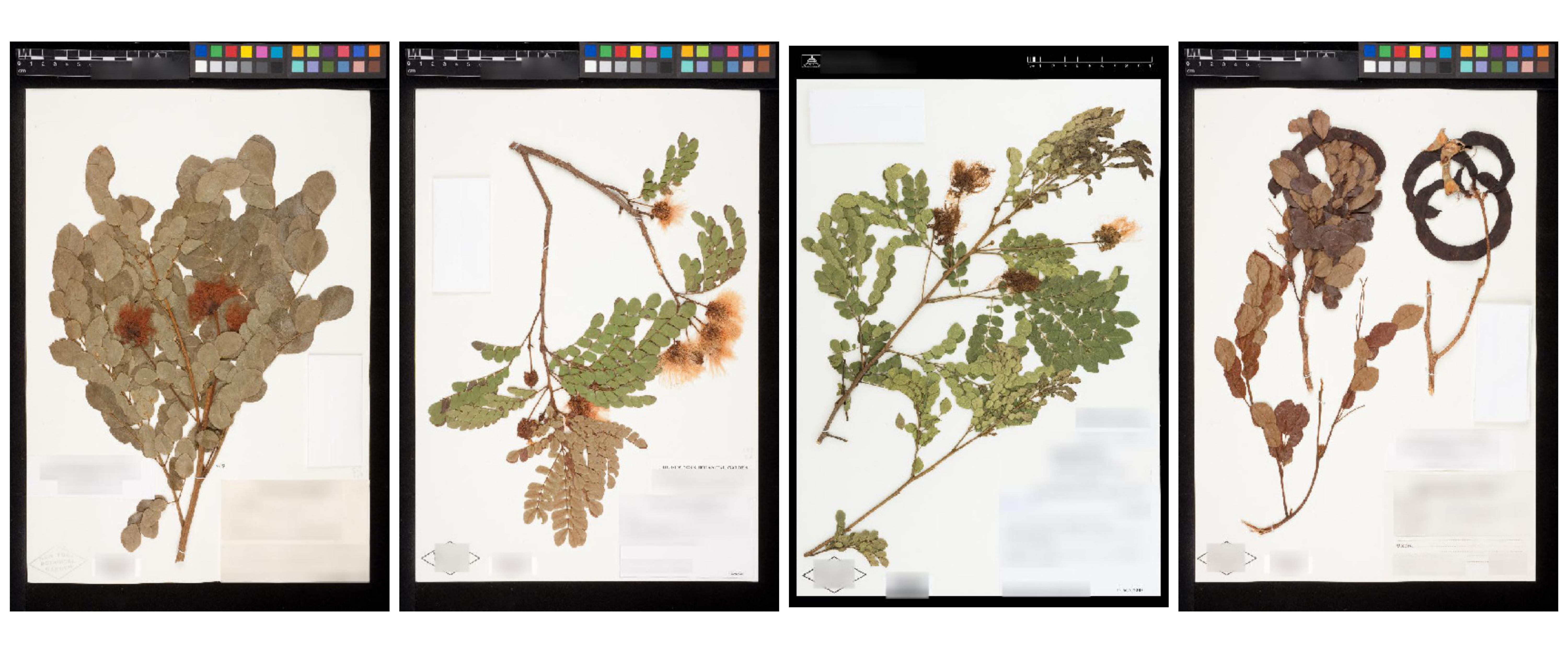}
\end{center}
   \caption{Example of visually different images corresponding to the same species: \textit{Abarema brachystachya} (DC.) Barneby \& J.W.Grimes (Fabaceae). The observed differences are primarily due to different reproductive stages: early flowering, late flowering, and fruit.}
\label{fig:sameSpecies}
\end{figure}

\begin{figure}[h]
\begin{center}
\includegraphics[width=0.95\linewidth]{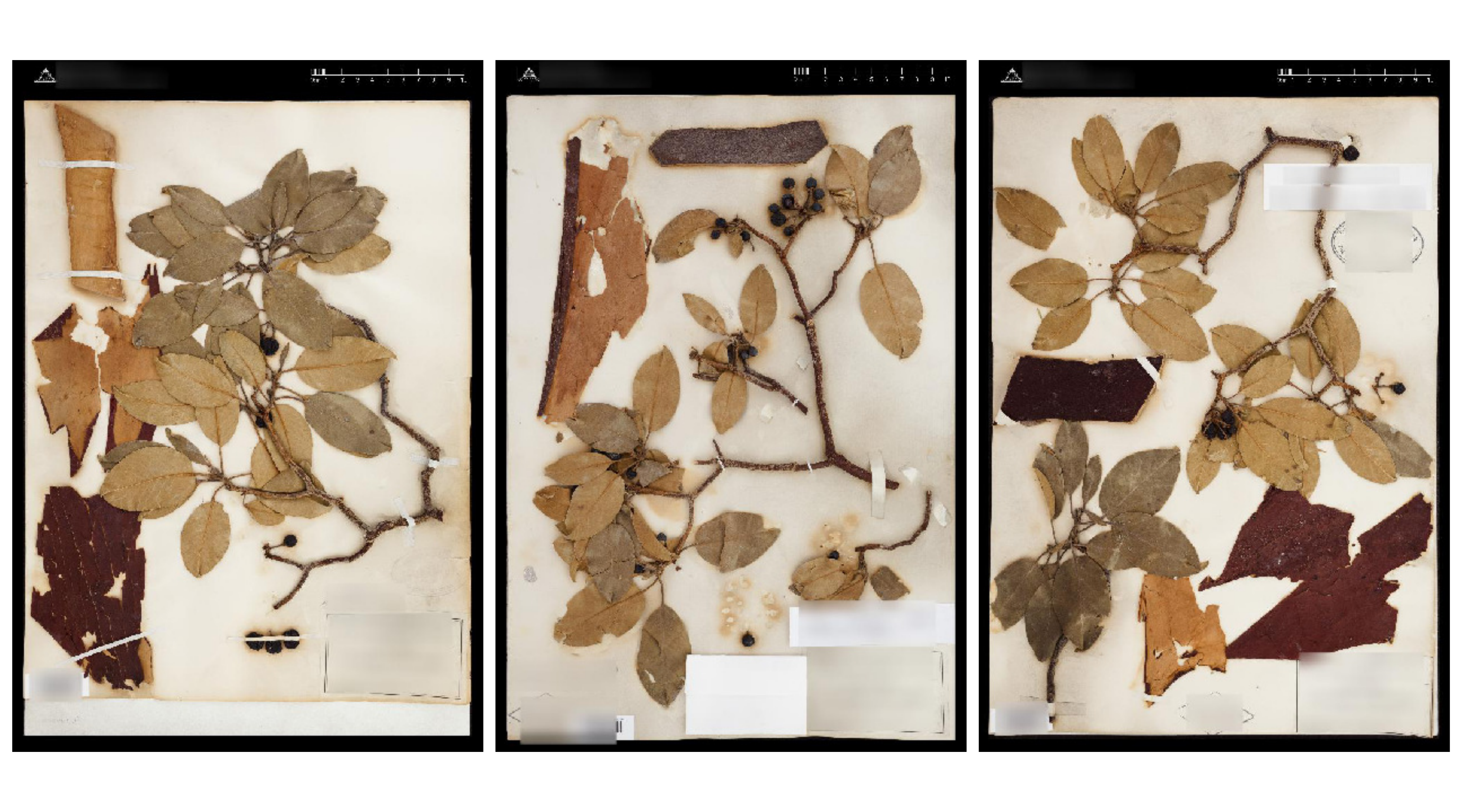}
\end{center}
   \caption{Different specimens of \textit{Arbutus xalapensis} Kunth (Ericaceae) made from the same individual plant at the same time by the same collector using the same pressing, drying, and mounting protocol.}
\label{fig:differentSpecimens}
\end{figure}

\begin{figure}[h]
\begin{center}
\includegraphics[width=0.95\linewidth]{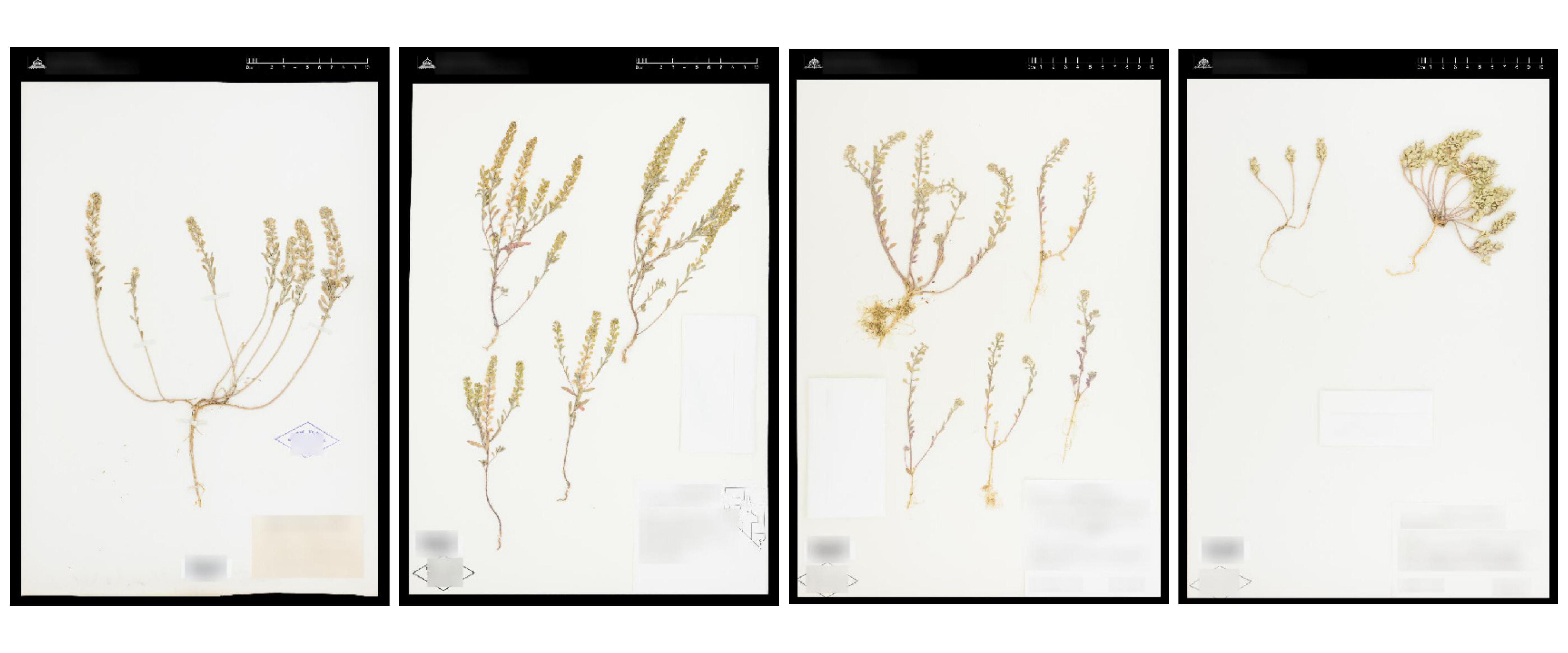}
\end{center}
   \caption{Example of visually similar images from different \textit{Alyssum} species (Brassicaceae): \textit{A.~alyssoides} (L.) L., \textit{A.~desertorum} Stapf, \textit{A.~simplex} Rudolphi, \textit{A.~szovitsianum} Fisch.~\& C.A.~Mey.}
\label{fig:differentSpecies}
\end{figure}

The Herbarium 2021 Half--Earth dataset is challenging due to multiple reasons. First, of course, due to its large imbalance (Fig.~\ref{fig:dist}), the imbalance factor for the dataset is 1,654.5.  Second, the variation within species (Figs.~\ref{fig:sameSpecies}) is high.  Herbarium specimens can capture plants at different growth--stages (e.g., juvenile versus adult) or with different sets of plant parts (e.g., leaves and flowers versus leaves and fruit; see Fig.~\ref{fig:sameSpecies}). In addition, the techniques used to press, dry, and mount specimens vary among collectors and collecting expeditions---these differences can change the appearance of specimens dramatically (e.g., collecting in alcohol often causes leaves to turn black). Arbitrary aesthetic decisions made while processing specimens can result in specimens that differ dramatically in appearance even though they are simply different parts of the same individual plant (Fig.~\ref{fig:differentSpecimens}). In a herbarium collection, every attempt to conserve dried specimens is made, but in practice older specimens become more fragile and suffer damage as they age leading to some specimens being less complete and more damaged than others. Third, the visual similarity among species can be high (Fig.~\ref{fig:differentSpecies}). Finally, the diagnostic morphological features that botanists use to identify species are often very small and thus require a model that is able to handle high--resolution images and can focus on specific details \cite{cope2012,waldchen2018}.
\subsection{Data Preprocessing}

In this section we give an overview of how we preprocessed the dataset. Figure \ref{fig:prepro} presents some example of herbarium sheets before and after the preprocessing steps. 

\subsubsection{Label Alignment}

Herbarium specimens of the same taxon may have been labeled in various ways due to differences in the interpretation of taxon circumscriptions, nomenclatural changes, and/or errors. For example, over time \textit{Pilosella piloselloides} (Vill.) Soják (Asteraceae) has been known by at least 526 different names \cite{LCVP}. 
To ameliorate this situation as much as possible, we have standardized the image labels to the Leipzig Catalogue of Vascular Plants (LCVP v1.0.2) \cite{LCVP}. Labels in our dataset have an LCVP status of either ``accepted'' or ``unresolved''. 

The data exported from the institutional databases were first processed to find labels that exactly matched LCVP. For labels that did not precisely match, we then searched for long unambiguous partial matches to LCVP: the label was shortened by removing the rightmost word and then we searched for a match that produced only one LCVP output taxon; if no match was found, this was repeated until the label contained only two words. Labels that still did not unambiguously match LCVP, were matched using tre-agrep \cite{WuManber1992} allowing an increasing amount of mismatch (10--30\% of label length; all weights were set to 1). Matches returned by tre-agrep were manually reviewed (8,430 labels passed manual review). Images with labels that could not be coerced into matching LCVP were excluded from the dataset (\textit{c}.~73 thousand images).

\subsubsection{Image Blurring}

Herbarium specimens always have a hand--written or printed label on the sheet (usually lower right--hand corner), which includes information about the name of the taxon, the geographic location where it was collected, the date of collection, and the person or team of people who collected it. In addition, annotation labels are often added to the specimen to correct or update information on the original label---these are sequentially added in the empty space above the original label. Specimens often also have institutional labels or stamps indicating the herbarium in which the specimen is archived and a barcode label corresponding to an institutional database entry. Specimens may also include field tags with identification numbers attached directly to the plant. Images usually include color and measurement scales as well as institutional logos. All of these labels can of course, help identify the specimen, thus we blurred this information in the dataset in order to force models to learn about the plants themselves rather than the label text. 

\begin{figure}[h]
\begin{center}

\includegraphics[width=0.95\linewidth]{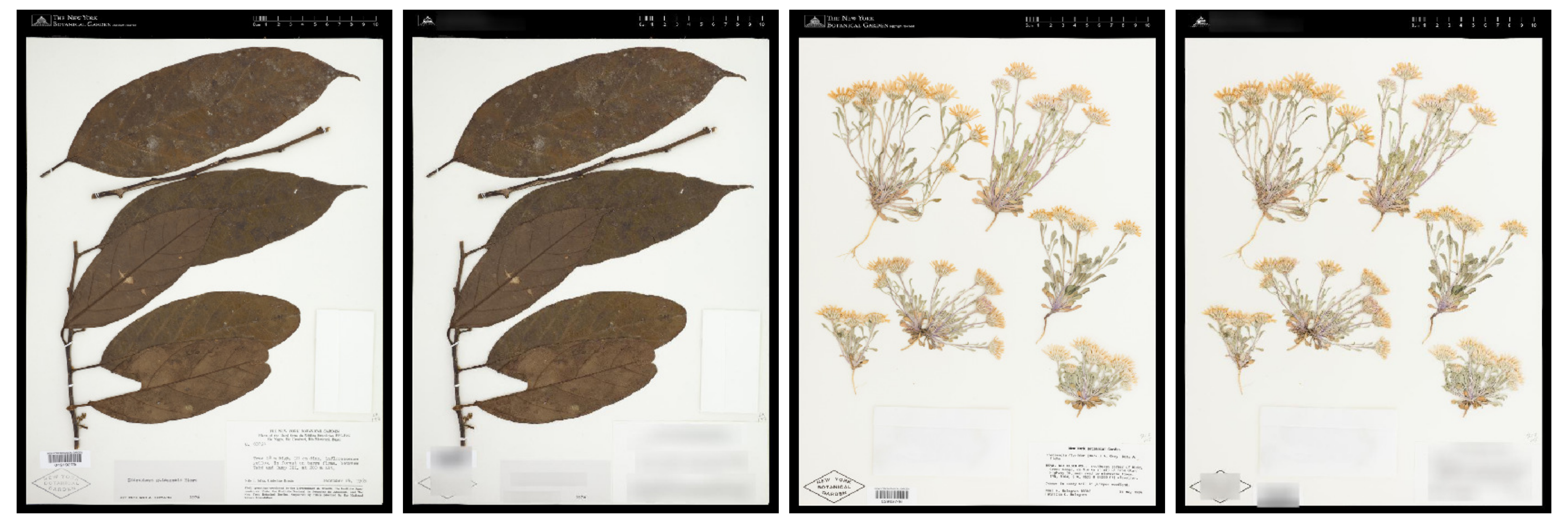} \\ 
\includegraphics[width=0.95\linewidth]{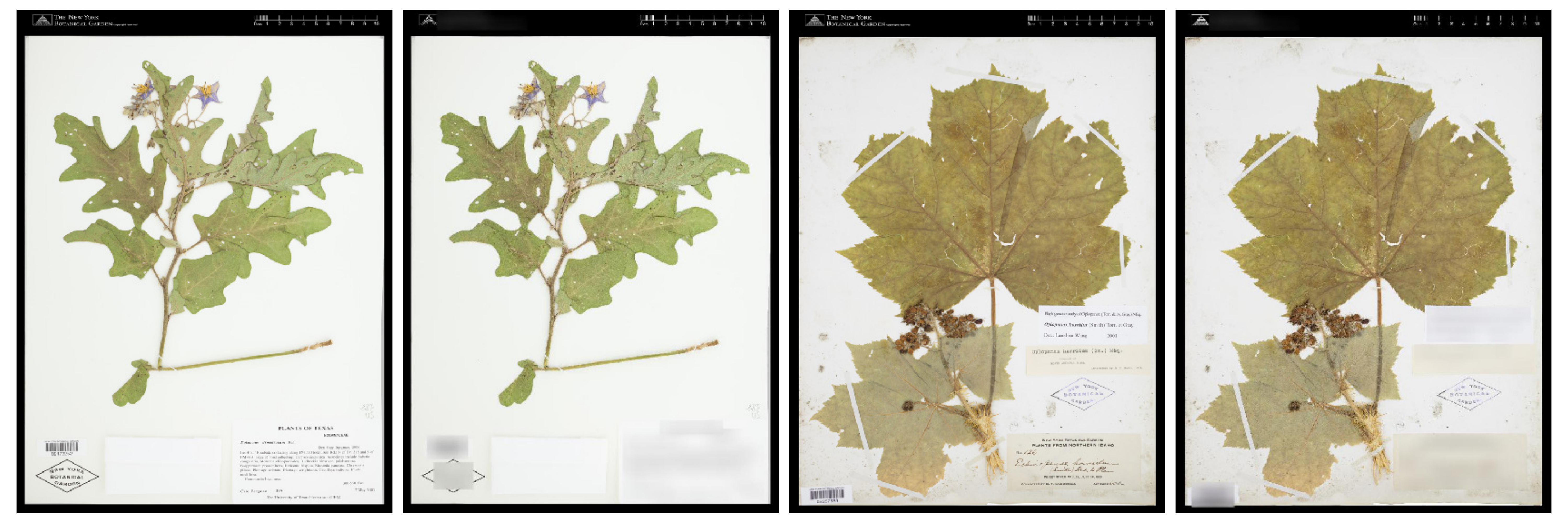} \\ 
\includegraphics[width=0.95\linewidth]{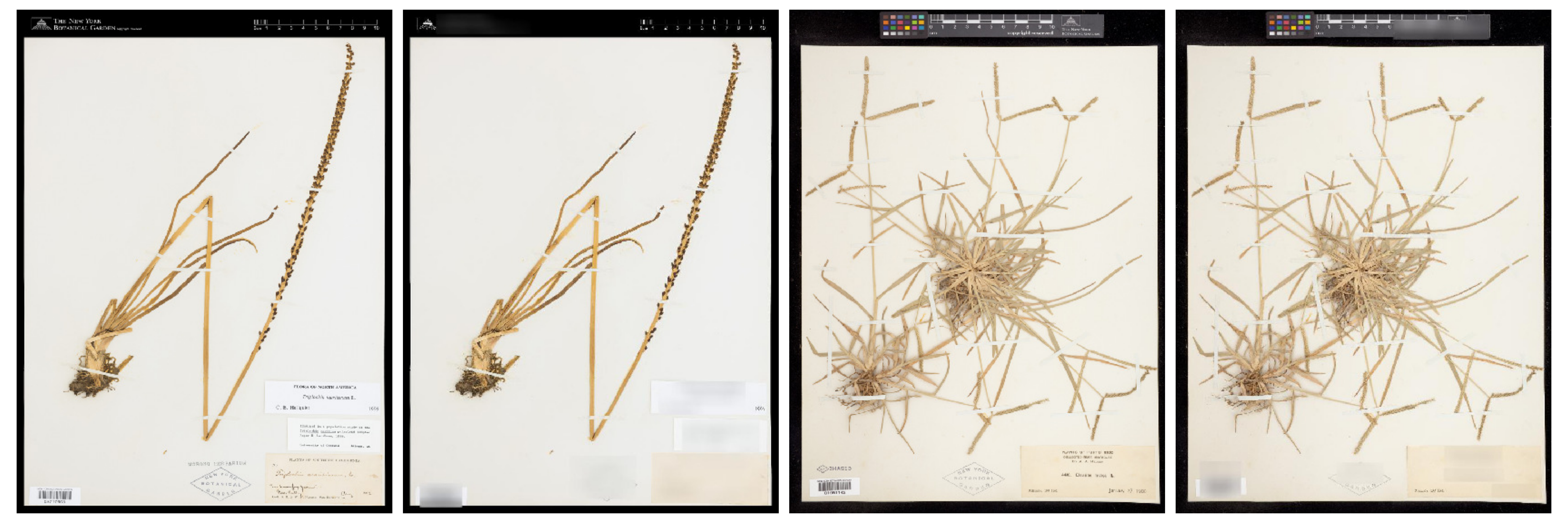} \\ 
\includegraphics[width=0.95\linewidth]{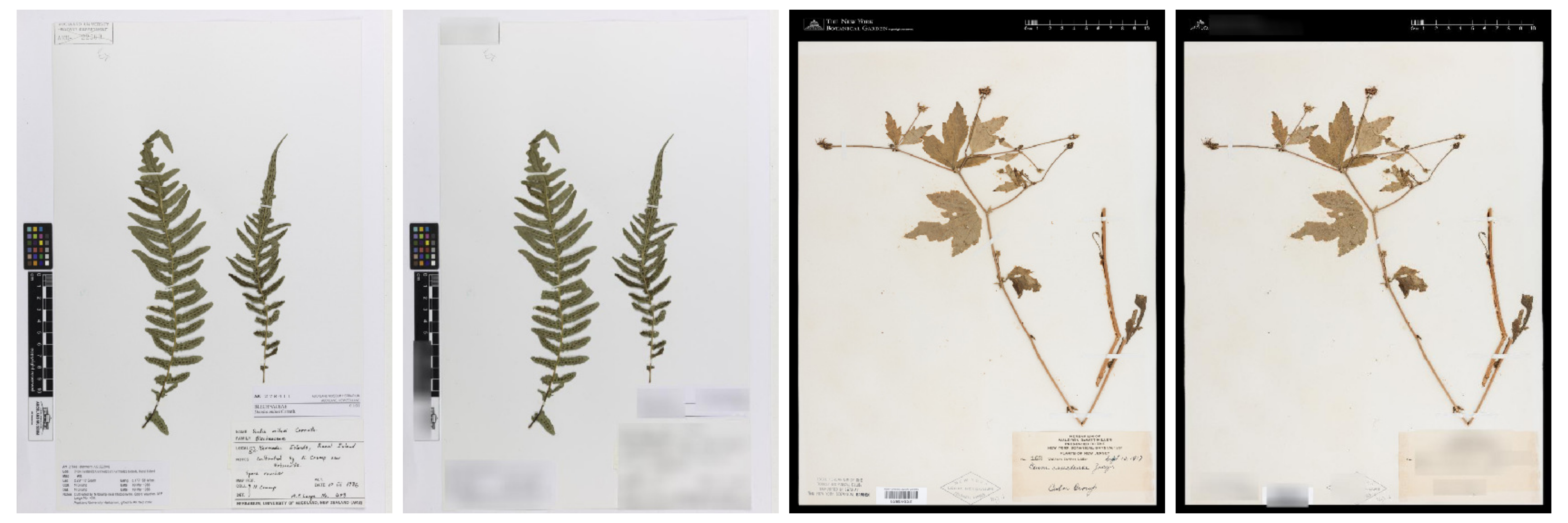} 

\caption{Example of images before and after preprocessing.}
\label{fig:prepro}
\end{center}
\end{figure}

To detect these labels we used a pretrained EAST text detection model \cite{EAST}. This model outputs bounding boxes around the detected text. We merged the bounding boxes that overlapped by a sufficient margin, and filtered out those that were too small. The resulting regions were then heavily blurred. We first applied a mean blur, then a single Gaussian blur with added noise, and finally used a smooth alpha map to blend into the original (Fig.~\ref{fig:prepro}). Finally, we excluded images from the dataset where more than 25\% of the image was blurred, as we found those to be, in most cases, wrong predictions from our text detection model. We deliberately chose to tune the text detection model to have a high specificity, in order to avoid unnecessarily blurring parts of the plant. Even though, this means that there are images where part of the labels are missed by our blurring algorithm.

\subsubsection{Image Resizing}

Herbarium sheets are digitized as very high--resolution images to preserve as much of the detail as possible. A common image size is around $6000 \times 4000$. This is very large even for networks that are designed to work with higher resolutions. We have resized all images in the dataset to a larger dimension of 1000 (while preserving the aspect ratio), in order to make the overall size of the dataset more accessible. 

\subsubsection{Dataset Split}

Our dataset contains images from 64,500 taxa at the species--level or below with 2,257,759 in the training set and 243,020 in the test set. The data has been split to obtain an approximately even number of images across taxa in the test set. In fact, we capped the maximum number of images per taxon to 10 for the test set. For taxa that have few images we did a 80\%/20\% split for training/test. Each category has a minimum of three images: at least one in the test set and two in the training set. 

\section{The Herbarium 2021 Half-Earth Challenge}

The Herbarium 2021 Half-Earth Challenge is a competition hosted on Kaggle as part of $8^{th}$ workshop for Fine--Grained Visual Categorization at CVPR 2021. This is the third iteration of the Herbarium Challenge, in this section we give a brief description of the previous challenges. 

\paragraph{The Herbarium 2019 Challenge} \cite{herbarium2019,herbarium_journal} focuses on the flowering plant family Melastomataceae. It contains 46,469 digitally imaged herbarium specimens representing 683 species. The Melastomataceae is a large family with 166 recognized genera and 5,892 species \cite{LCVP}. The overlap with the iNaturalist 2018 challenge dataset \cite{inat2017} is only 2 out of the 683 species in the Herbarium 2019 dataset. 

\paragraph{The Herbarium 2020 Challenge} dataset contains over 1M images representing over 32,000 plant species. This challenge focuses on vascular land plants of the Americas. 

\section{Baseline and Evaluation Metric}

As a simple baseline we have trained a ResNet-50 \cite{resnet} for 10 epochs. We split the training set in a stratified manner to create a hold-out validation set, thus the baseline was trained on 80\% of the full training set. We used a balanced sampling strategy, so as to mitigate the impact of the imbalance on the classifier. We resized the images to $256\times256$  and used some standard data augmentations (small rotations, horizontal flips, color-jitter and finally center-crop to $224\times224$). We initialised the model with weights pretrained on ImageNet \cite{imagenet}. Finally we trained the model using the standard cross-entropy loss, a batch size of $32$, a stochastic gradient descent with a learning rate of $1\cdot10^{-3}$ which is further reduced when a plateau is reached and a momentum factor of $0.9$.

The evaluation metric for the Herbarium 2021 Half-Earth Challenge is the $F_1$ score:
\begin{equation}
F_1=2\frac{\mathrm{Pre}\cdot  \mathrm{Rec}}{\mathrm{Pre} + \mathrm{Rec}}~,
\label{eq:f1}    
\end{equation}
where $\mathrm{Pre}$ denotes the precision and $\mathrm{Rec}$ the recall. Our baseline achieves an $F_1$ score of $0.46$ on the private test set of the competition. For comparison, the first place solution of the competition achieved an $F_1$ score of $0.76$ on the private test set \footnote{Results still unverified at the moment of writing.}.

\section{Conclusion}

We presented the Herbarium Half-Earth dataset to enable the development of better automatic taxon recognition models. The development of models to automatically identify specimens will reduce the bottleneck of species identification and has the potential to advance biodiversity research at an unprecedented rate.

In the future, we would like to expand the dataset to include specimens collected world-wide. There are more than 35 million digitized specimens in electronic databases representing more than 80\% of the known vascular plant diversity. 

\paragraph{Acknowledgments}
We would like to thank our colleagues at the New York Botanical Garden (Damon Little, Kimberly Watson, Barbara Ambrose) and Kiat Chuan Tan from Google for their generous support in making this challenge possible. We are grateful for the participation of our collaborators at Auckland War Memorial Museum (Dhahara Ranatunga, Yumiko Baba), Bishop Museum (Melissa Tulig), Queensland Herbarium at Brisbane Botanic Gardens (Gillian Brown, Gordon Guymer, Andrew Franks), Naturalis Biodiversity Center (Jan Wieringa) and for contributing images to this dataset.

{\small
\bibliographystyle{./aux/ieee_fullname.bst}
\bibliography{bibliography.bib}
}

\end{document}